\pdfoutput=1
\documentclass[11pt]{article}

\usepackage[]{acl}

\usepackage{times}
\usepackage{latexsym}

\usepackage[T1]{fontenc}
\usepackage{graphicx}
\usepackage{caption}
\usepackage{subcaption}
\usepackage{xcolor}
\definecolor{capri}{rgb}{0.0, 0.75, 1.0}
\definecolor{chromeyellow}{rgb}{1.0, 0.65, 0.0}
\definecolor{carminepink}{rgb}{0.92, 0.3, 0.26}
\definecolor{cambridgeblue}{rgb}{0.64, 0.76, 0.68}
\definecolor{denim}{rgb}{0.08, 0.38, 0.74}
\definecolor{bluegray}{rgb}{0.4, 0.6, 0.8}
\usepackage[utf8]{inputenc}

\usepackage{microtype}

\usepackage{inconsolata}
\usepackage{amsmath}
\usepackage{booktabs}
\usepackage{microtype}
\usepackage{graphicx}
\usepackage{amsmath}
\usepackage{tikz}
\usepackage{multirow}
\usepackage{overpic}

\usepackage{amssymb}% http://ctan.org/pkg/amssymb
\usepackage{pifont}% http://ctan.org/pkg/pifont

\newcommand{\pk}[1]{{\textcolor{red}{~(PK: #1)}}}
\title{
NavHint: Vision and Language Navigation Agent with a Hint Generator}

\author{Yue Zhang \\
  Michigan State University \\
  \texttt{zhan1624@msu.edu} \\\And
  Quan Guo \\
  Sichuan University \\
  \texttt{guoquan@scu.edu.cn}\\\And
  Parisa Kordjamshidi \\
  Michigan State University \\
  \texttt{kordjams@msu.edu}
  }

\begin{document}
\maketitle
\maketitle
\begin{abstract}

Existing work on vision and language navigation mainly relies on navigation-related losses to establish the connection between vision and language modalities, neglecting aspects of helping the navigation agent build a deep understanding of the visual environment.
In our work, we provide indirect supervision to the navigation agent through a hint generator that provides detailed visual descriptions.
The hint generator assists the navigation agent in developing a global understanding of the visual environment. It directs the agent's attention toward related navigation details, including the relevant sub-instruction, potential challenges in recognition and ambiguities in grounding, and the targeted viewpoint description. 
To train the hint generator, we construct a synthetic dataset based on landmarks in the instructions and visible and distinctive objects in the visual environment.
We evaluate our method on the R2R and R4R datasets and achieve state-of-the-art on several metrics. 
The experimental results demonstrate that generating hints not only enhances the navigation performance but also helps improve the interpretability of the agent's actions.

\end{abstract}

\section{Introduction}
% VLN task
%Explanation abilities are critical for effective interactions between a navigation agent and human beings. 
In many real-world applications, it is a crucial skill for an intelligent agent to perceive the visual environment and interact with humans using natural language. % understanding human's natural language and perceiving the visual environment.
%An intelligent agent that can understand human's natural language while perceiving the visual environment and take actions is one of fundamental goals in artificial intelligence.
The Vision and Language Navigation (VLN) task~\cite{anderson2018vision} is one of the popular problems in this direction that has attracted significant attention from computer vision, natural language processing, and robotic communities~\cite{li2022envedit, fried2018speaker, francis2022core}.
% Explanation abilities are critical for effective interactions between a navigation agent and human beings. 
% However, e
% use different vision representations and augment vision representations, language community by improving the grounding ability in both objects and spatial relations.
%todo: citations

\begin{figure}
    \centering\includegraphics[width=1.0\linewidth]{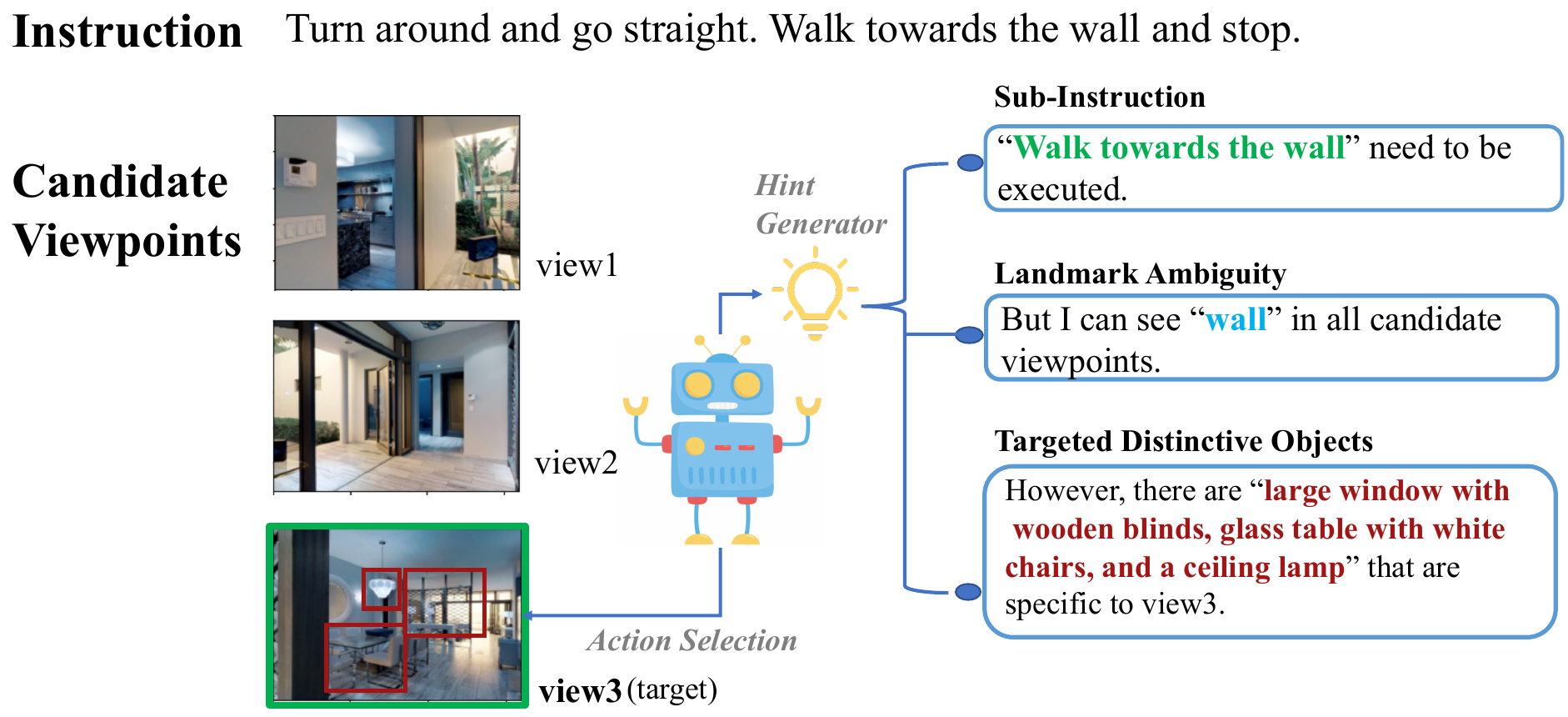}
    \caption{Given the instruction and three candidate viewpoints, the navigation agent with the assistance of the hint generator, produces descriptions of the visual environment with three key elements: sub-instruction, landmark ambiguity and targeted distinctive objects.\vspace{-4mm}}
    \label{fig:motivation1}
\end{figure}

% Motivation
With the increasing popularity of the VLN task, many neural navigation models~\cite{hong2020recurrent,chen2021history, hao2020towards} have been proposed.
One line of research is to strengthen the connection of the vision and language modalities~\cite{ma2019self, hong2020language,li2021improving}.
However, the majority of these efforts learn the connection mainly supervised by navigation performance,
such as the distance to the destination, the orientation selection (heading and elevation), and the similarity between the given instruction and the trajectory.
While this helps teach the agent to navigate, it does not directly enforce learning comprehensive textual and visual semantics.
In fact, learning visual semantics in the environment is crucial not only for successfully completing navigation tasks but also for the effective communication with humans.
For instance, the navigation agent should correctly locate the navigation progress based on the current visual views.
Moreover, the navigation agent needs to adopt a global perspective of the environment to investigate whether the navigable viewpoints include the relevant landmarks or whether the instruction is ambiguous. In any case, the agent should be able to describe its targeted viewpoint.
Expecting the navigation agent to obtain the above understanding solely through navigation-related signals is challenging, and the intermediate guidance is necessary.

To this end, we introduce a hint generator for the VLN agent (NavHint), aiming to generate visual descriptions that serve as indirect supervision to help the navigation agent obtain a better understanding of the visual environment (as depicted in Fig.~\ref{fig:motivation1}). 
When the agent navigates at each step, the hint generator concurrently produces visual descriptions that are consistent with the agent's action decision. 
The hints are designed based on the rationale underlying the navigation process, including three aspects:
\textit{Sub-instruction}, \textit{Landmark Ambiguity} and \textit{Targeted Distinctive Objects}.
Specifically, at each navigation step, \textbf{first}, the hint generator encourages the agent to report its navigation progress by specifying which part of the sub-instruction it is executing based on the current visual environment. As depicted in ~Fig.~\ref{fig:motivation1}, the sub-instruction ``walk towards the wall'' needs to be executed.
\textbf{Second}, the hint generator directs the agent to have a global view of the entire environment and recognize the landmarks mentioned in the instruction from all candidate viewpoints.
The agent is tasked with identifying potential challenges by assessing the visibility of the landmarks and comparing the landmarks shared among viewpoints. For instance, in the given example, the landmark "wall" is ambiguous as it appears in multiple views.
\textbf{Third}, in scenarios where challenges exist, the hint generator guides the agent in describing the distinctive visual objects that only appear in the targeted viewpoint, such as "\textit{large window with wooden blinds}" in view3 in Fig~\ref{fig:motivation1}. This aids the agent in deeply looking into the details of its selected viewpoint while globally comparing it to other candidates.

The hint generator is designed as a Transformer-based decoder that leverages visual output from the navigation agent to produce corresponding hints. 
% This hint generator have a high-level adaptability, in that it functions as a plug-in module for any VLN agent. 
This hint generator can be plugged into any VLN agent as a language model conditioned on the VLN models.
To train the hint generator, we propose a synthetic navigation hint dataset based on Room2Room (R2R)~\cite{anderson2018vision} dataset.
Our dataset provides hints for each step of the trajectory in the R2R dataset. Each hint description includes sub-instruction, landmark ambiguity, and targeted distinctive objects introduced above. 
The dataset serves as an extra supervision to train the navigation agent and the hint generator jointly.
Besides, our constructed dataset can be utilized to explicitly analyze the navigation agent's grounding ability by assessing the quality of generated hints.

In summary, our contributions are as follows:\\
1. We leverage a language model conditioned on the VLN models to design a hint generator that can be plugged into any VLN agent. This hint generator helps the agent develop a comprehensive understanding of the visual environment. \\
2. We construct a synthetic hint dataset to provide the agent with visual descriptions at each navigation step. The dataset serves as an indirect supervision for jointly training the navigation agent and the hint generator.\\
3. We show that the hint generation improves the agent's navigation performance on the R2R and R4R datasets. 
% Also, detailed analysis shows generated hints improves interpretability of the agent's decisions.
We also provide a detailed analysis of the agent's grounding ability by examining the quality of the generated hints, thereby improving the interpretability of the agent's decisions.

% \begin{figure*}
%     \centering
%     \begin{overpic}[width=1.0\linewidth]{images/data1.pdf}  
%     %% caption
%     \put(30,-0.5){\small{(a)}}
%     \put(80,-0.5){\small{(b)}}
%     \end{overpic}
%     % \includegraphics[width=1.0\linewidth]{images/data.pdf}
%     \caption{Navigation Hint Dataset. (a) An example of a navigation hints with the landmark ambiguity of ``\textit{Missing Landmarks}''. The sub-instruction is``walk into the hallway''(\textcolor{chromeyellow}{\rule{0.4cm}{0.25cm}}), and the landmark ``hallway'' (\textcolor{capri}{\rule{0.4cm}{0.25cm}}) in the instruction is observed in the view1 rather than target view3, which can potentially mislead the navigation agent. The target distinctive objects "wooden dining table" and "marble countertop."(\textcolor{carminepink}{\rule{0.4cm}{0.25cm}}) are then provided. "Blue walls" (\textcolor{bluegray}{\rule{0.4cm}{0.25cm}}) is non-distinctive as it appears in both view2 and view3. 
%     (b) Statistics of different categories of landmark ambiguity. \vspace{-4mm}}
%     \label{fig:Reasoning Dataset Construction}
% \end{figure*}

\section{Related Work}

\textbf{Navigation Instruction Following} \newcite{anderson2018vision} first extended the instruction following to the photo-realistic simulated environments. 
Subsequent studies have emerged with an emphasis on enhancing navigation performance through multi-modal learning~\cite{hong2020language, wang2023scaling, zhang2022explicit, an2021neighbor, zhang2021towards}, map representation learning~\cite{hong2023learning, chen2022weakly,an2023etpnav}, or graph-based explorations~\cite{zhu2021soon,wang2021structured, chen2022think}. 
One line of effort has been to provide auxiliary reasoning tasks or pre-training proxy tasks to guide the navigation agent to learn textual and visual representations~\cite{zhu2020vision, chen2021history, hao2020towards, qiao2022hop, zhang2022lovis}. AuxRN~\cite{zhu2020vision} proposes four auxiliary reasoning tasks to gain knowledge of the navigation map and the consequences of actions. 
However, most of those methods acquire the textual and visual semantics from a wayfinding perspective during navigation, which may be insufficient 
% to help 
for agents to understand the visual environment comprehensively. We address this issue with our proposed hint generator that offers visual descriptions 
% as indirect supervision 
to guide the navigation agent in learning visual semantics.

% However, 

% However, most of them mainly focus on the navigation agent's wayfinding performance and ignore to explore the deeper understanding of the visual environment.

% its explanation ability. While explicit semantic modeling attempts to enhance grounding ability, it relies on representation learning, making it challenging to explain the agent's actions.
% For example, from the vision side, \newcite{zhang2020diagnosing} investigates how different visual representations influence navigation performance. EnvEdit~\cite{li2022envedit} augments the visual images by editing image styles. From the language side, \newcite{hong2020sub} segments the long instruction into
% sub-instructions and annotates their corresponding
% trajectories.
% \newcite{li2021improving} propose to utilize syntax information like dependencies and phrase structures to improve cross-modal alignment.
% One line of such effort is to model the semantic structure of both text and vision modalities~\cite{hong2020language, zhang2022explicit, an2021neighbor, zhang2021towards}. For example, RelGraph~\cite{hong2020language} builds an implicit language-visual entity relation graph to learn the connection between the text and vision. 
%The second line of effort is to learn vision and language cross-modality representations~\cite{hao2020towards, hong2020recurrent, chen2021history}. \newcite{hong2020recurrent} directly apply Transformer for the navigation task and design a recurrent state 

\noindent\textbf{Language-Capable VLN Agent} A few studies attempt to design language-capable VLN agents to improve the agent's grounding ability. 
Most of the work encourages the navigation agent to reproduce the original instruction. For example,
LANA~\cite{wang2023lana} devises an agent that executes human-written navigation commands and provides route descriptions. 
Similarly, one of the tasks in AuxRN~\cite{zhu2020vision} is to retell the trajectory.
However, these approaches have limitations because the original instruction can sometimes be inaccurate and confusing, as suggested in the VLN-Trans~\cite{zhang2023vln}. Forcing the agent to reproduce the same instruction in such cases can undermine the agent's grounding ability.
Instead of only focusing on the original instruction, our proposed hint generator produces visual descriptions from a global perspective, thereby enhancing the agent's understanding of the visual environment and improving its grounding ability.

% and does not contribute to the agent's own understanding of the environment.
%Our approach assists the agent in understanding the visual environment and enhancing the agent's grounding ability.

% pinpointing potential challenges and distinctive objects in the targeted viewpoint.

% Although VLN-Trans introduces a translator to convert the original instructions into sub-instruction based on the agent's visualization ability, their method is implicitly sub-instruction representation learning rather than generating explicit natural language instructions, which still does not improve the interpretability of the model.

\section{Method}

\begin{figure}
    \centering
    \begin{overpic}[width=1.0\linewidth]{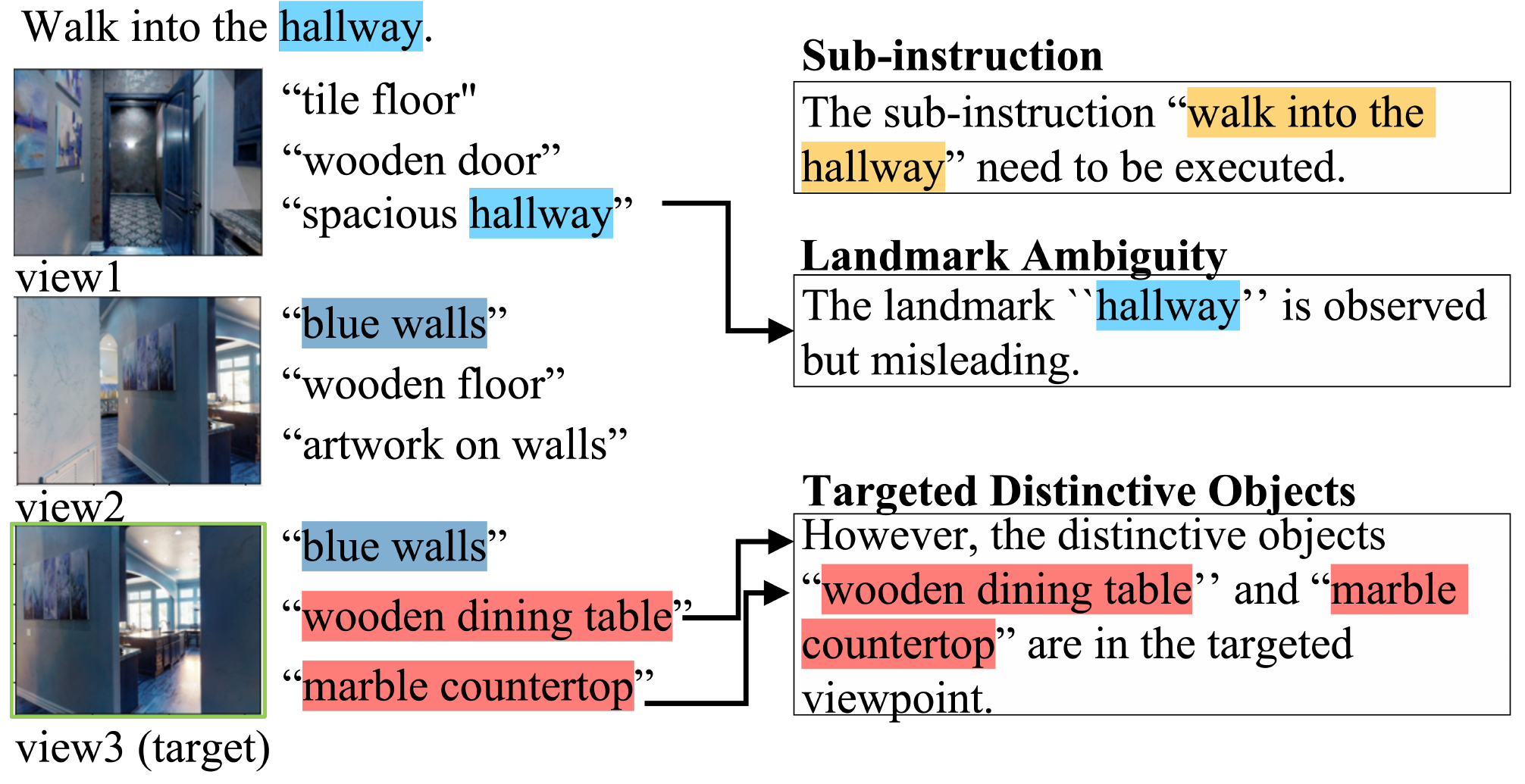}
    \end{overpic}
    \caption{Navigation Hint Dataset. An example of a navigation hints with the landmark ambiguity of ``\textit{Missing Landmarks}''. The sub-instruction is``walk into the hallway''(\textcolor{chromeyellow}{\rule{0.4cm}{0.25cm}}), and the landmark ``hallway'' (\textcolor{capri}{\rule{0.4cm}{0.25cm}}) in the instruction is observed in the view1 rather than target view3, which can potentially mislead the navigation agent. The target distinctive objects "wooden dining table" and "marble countertop."(\textcolor{carminepink}{\rule{0.4cm}{0.25cm}}) are then provided. "Blue walls" (\textcolor{bluegray}{\rule{0.4cm}{0.25cm}}) is non-distinctive as it appears in both view2 and view3. \vspace{-4mm}}
    \label{fig:Reasoning Dataset Construction}
\end{figure}

In the VLN problem setting, the agent is given a natural language instruction, denoted as $W = \{w_1, w_2, \cdots, w_l\}$, $l$ is the length of the sentence. At each navigation step, the agent perceives panoramic views with $36$~\footnote{$12$ headings and $3$ elevations with $30$-degree intervals.} discrete images. There are $n$ candidate viewpoints that can be navigated to, denoted as $I=\{I_1, I_2,\cdots, I_n\}$. This task aims to generate a trajectory following the given instruction. 
% The navigation process can be terminated when the agent selects the same action or the maximum navigation steps are reached.
%Figure~\ref{fig:Architecture} shows an overall picture of our method that
In the following section, we first present our constructed navigation hint dataset. Then, we introduce the hint generator. The navigation hint dataset is used to train the navigation agent and the hint generator jointly.

% that can be plugged into the navigation agent, aiming to improve the navigation agent's understanding of the visual environment. 
% 
%We describe the details of our method in the following sections.

% \pk{Maybe name the dataset something like "waypoint explanations", "VLN explanation" "route commentary" or "scene annotations", "scene explanation dataset" you can ask chatgpt for better words.}

\subsection{Navigation Hint Dataset}
\label{hintdataset}
%In this section, we introduce our approach for constructing  the VLN explanation dataset. 
% \pk{We construct an explanation dataset aiming for VLN grounding explanation based on the existing VLN dataset automatically.: We construct an explanation dataset automatically. The goal is to use it for training the agent to generate detailed explanations of their viewpoints and actions that requires their in-depth understanding of the environment and grounding language in vision.}

% \pk{AS WE discussed in the meeting, I think you need to emphasize the fact that training the agent for explanation generation enforces the agent to have a better understanding of the environment and therefore results in a better navigation performance at the end. This understanding includes pinpointing the actual part of the instruction that needs to be executed;  understanding the commonalities of multiple views in terms of shared landmarks that may cause ambiguity; and finding their distinctive landmarks that help finding the target view.}

The purpose of constructing the navigation hint dataset is to provide supervision for the hint generator to generate detailed visual description. 
% The dataset further helps the navigation agent acquire the ability to ground language into visual observations and obtain a deep understanding of the visual environment. 
The navigation hint dataset is automatically generated based on instruction and trajectory pairs from the R2R dataset~\cite{anderson2018vision}. 
For every step of the trajectory, we 
provide hints that mainly include three key elements, as described below.

% \pk{. It includes explanations related to multiple aspects of navigation and details of visual observations } as we describe below.
% Specifically, our explanation dataset helps the navigation agent from the following three aspects: pinpointing a part of the instruction that needs to be executed, understanding the commonalities of multiple views in terms of shared landmarks that may cause ambiguity, and finding the distinctive objects in the target view that help differentiate it from other viewpoints.
\begin{figure}
    \centering
    \begin{overpic}[width=0.85\linewidth]{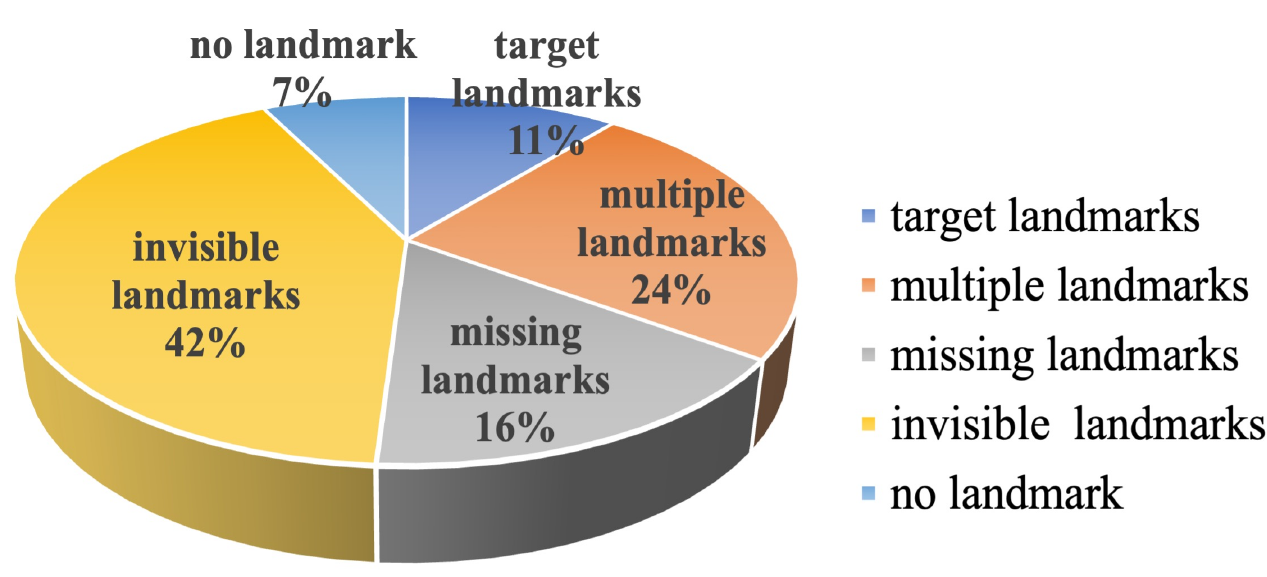}  
    \end{overpic}
    \caption{Statistics of different categories of landmark ambiguity. \vspace{-4mm}}
    \label{fig:Statistics of different categories}
\end{figure}

\begin{table*}[]
    \centering
    \small
    \resizebox{1.0\textwidth}{!}{
    \begin{tabular}{c|c|c}
    \hline
        \textbf{Ambiguity Category} & \textbf{Description} & \textbf{Hints}\\
        \hline
        Target Landmarks & Landmarks only appear in the target. & The \{landmarks\} are observed.\\
        \hline
        Multiple Landmarks & Landmarks are visible in multiple viewpoints including the target viewpoint. & The \{landmarks\} are observed in multiple viewpoints.\\
        \hline
        Missing Landmarks & Landmarks are visible in other viewpoints except for the target viewpoint. &  The\{landmarks\} are misleading.\\
        \hline
        Invisible Landmark & Landmarks are not visible in all viewpoints & The\{landmarks\} are not observed.\\
        \hline
        No Landmarks & No landmarks in sub-instruction. (\textit{e.g.} ``\textit{make a right turn}'', ``\textit{turn left}'', and ``\textit{go straight}'') &  $\varnothing$ \\
        \hline
    \end{tabular}
    }
    \caption{Landmark Ambiguity. The col\#1 and col\#2 show the categories of landmark ambiguity and the corresponding descriptions. The col\#3 shows the template for generating the hint for each category.\vspace{-5mm}}
    \label{tab:The category of Challenges}
\end{table*}

% \subsubsection{Explanation Content}

%In the following section, we will introduce how we establish each component.\\
% The aim of our explanation is to used as supervision for the navigation agent to improve its grounding ability.
% The explanation contains progress grounding, aiding in understanding the navigation progress; commonality grounding, providing clarification on navigation challenges; and target distinction grounding, facilitating a better understanding of how the selected view differs from other viewpoints. In the following sections, we introduce how we obtain each part.\\
% \pk{I am not in favor of repeating "grounding" for the three items, it looks superficial.}
% \pk{please note, for the following description you need to explain 1) the generation process of each part such as your template etc and 2) the motivation for placing that part into the explanation.}\\
\noindent\textbf{Sub-instruction} is the first part of the hint that pinpoints to the relevant part of the instruction (sub-instruction) to be processed at the current step.
We obtain the sub-instructions and their corresponding viewpoints from the FGR2R~\cite{hong2020sub} dataset, which provides human annotations of sub-instructions and the aligned viewpoints.

After obtaining the sub-instruction at each step, we insert it into our hint template, which is "\textit{The \{sub-instruction\} needs to be executed.}". 
Guiding the navigation agent to detect the related sub-instruction at each step is crucial since it effectively assists the agent in tracking its navigation progress.
\\
\noindent\textbf{Landmark Ambiguity} is the second part of the hint that describes the commonalities across multiple views that can result in ambiguity during navigation. This part of hint is achieved by examining the shared landmarks mentioned in the instruction among the candidate viewpoints.
% For generating this part, the model has to ground the mentioned landmarks in the instruction into the visual objects in all candidate viewpoints and identify the source of ambiguity.
% \pk{is to ground the mentioned landmarks in the instruction to the visual objects in all candidate viewpoints while also identifying the commonalities across multiple views that may lead to ambiguity in navigation.:

To automatically generate this part of the hint for building the dataset,
we first use spaCy\footnote{https://spacy.io/} to extract noun phrases from sub-instruction and use them as landmarks.
Then, we extract visual objects in each candidate viewpoint using MiniGPT-4~\cite{zhu2023minigpt}\footnote{https://minigpt-4.github.io/} with a two-step textual prompting.
We choose visual objects generated by MiniGPT-4 instead of Matterport3D object annotations because Matterport3D objects are pretty limited, with only 40 object categories like ``\textit{doors}'', ``\textit{walls}'', and ``\textit{floors}''. These generic objects are not sufficient for resolving landmark ambiguity. Moreover, the absence of attribute annotations in Matterport3D poses a challenge for landmark disambiguation, such as the differences between ``\textit{wooden table}'' and ``\textit{glass table}''. In contrast, MiniGPT-4 can generate such detailed attribute descriptions.
Specifically, for each candidate viewpoint, we feed MiniGPT-4 with the viewpoint image, asking ``\textit{Describe the details of the image.}'' and then ``\textit{List the objects in the image}''.
The generated text is in free form, and we post-process it to retrieve a list of extracted object descriptions.
After obtaining textual landmark names and visual objects, we examine the shared landmarks among the candidate viewpoints.
The presence of shared landmarks can pose ambiguity for the navigation agent. We categorize the ambiguity into: \textit{Target Landmarks}, \textit{Multiple Landmarks}, \textit{Missing Landmarks}, \textit{Invisible Landmarks} and \textit{No Landmark}.
and their descriptions are in Table~\ref{tab:The category of Challenges}.
Fig.~\ref{fig:Statistics of different categories} shows the statistics of ambiguity of our navigation hint dataset. Most cases are ``\textit{Invisible Landmarks}'' or ``\textit{Multiple Landmarks}", which is consistent with the argument in VLN-trans~\cite{zhang2023vln} that invisible and non-distinctive landmarks cause issues for the navigation agent in following instructions.
% The category of `\textit{No Landmark}'' indicates the cases that the sub-instructions do not contain any landmark information, such as ``\textit{make a right turn}'', ``\textit{turn left}'', and ``\textit{go straight}''.

After identifying the category of landmark ambiguity, we construct this part of the hint using the corresponding templates in col \#3 of Table~\ref{tab:The category of Challenges}. Identifying landmark ambiguity requires the navigation agent to ground the mentioned landmark names in the instruction to the visual objects in all candidate viewpoints. 
Guiding the navigation agent to identify such detailed ambiguities can help enhance its understanding of the connection between the instruction and the entire visual environment.

\noindent\textbf{Targeted Distinctive Objects} is the third part of the hint that describes the distinctive visual objects specific to the targeted view.
% VLN-Trans~\cite{zhang2023vln} highlights the instances that non-distinctive landmarks is one of main issues that can cause the confusion to the navigation agents.
% However, they do not provide quantitative evidence how does this issue influence the navigation performance. 
% Then we collect $60$ examples belonging to the challenge of "multi-viewpoints" and manually check two aspects. The first is
% whether the issue of non-distinctive landmarks truly happen in those examples. 
% The second is how the navigation performance is influences.
% The results shows that non-distinctive happen in $56/60$ examples, and only $20/60$ example can navigate to the correct target. In such a case, the distinctive objects in the selected viewpoint plays a crucial role in guiding the agent towards the correct target.
The agent should be able to justify its decision by describing the distinction of the targeted view.
We follow the approach of obtaining distinctive objects in the VLN-Trans~\cite{zhang2023vln} that compares the visual objects in the targeted and other candidate viewpoints. 
The distinctive objects are the ones that exclusively appear in the targeted viewpoint and do not appear in other views.

The hint template for targeted distinctive objects is ``\textit{However,  \{the comma-separated list of distinctive object names\} are in the targeted view.}''. We use $3$ distinctive objects at most.
If the cases belong to the challenge of ``\textit{Target Landmark}'', there is no need to provide extra distinctive objects since the landmark is already exclusive to the targeted viewpoint.
Describing distinctive objects is important 
% for the agent 
to obtain a global understanding of the visual environment by highlighting the differences between the targeted viewpoint and other candidate viewpoints.

% Also for the challenge of ``\textit{No landmark}'', our explanations will present distinctive visual objects in the target viewpoint, thereby providing additional visual information.
We collect hint for  each step of trajectory to construct our navigation hint dataset.
% Fig.~\ref{fig:Reasoning Dataset Construction} has shown an example with landmark ambiguity of ``\textit{Missing Visible}'' and how we construct each part for the explanations. 
% We also provide statistics of the dataset in Appendix~\ref{appendix: data statistics.}.
More details are in Appendix~\ref{appendix: data statistics.}.

\begin{figure*}
    \centering   \includegraphics[width=1.0\linewidth]{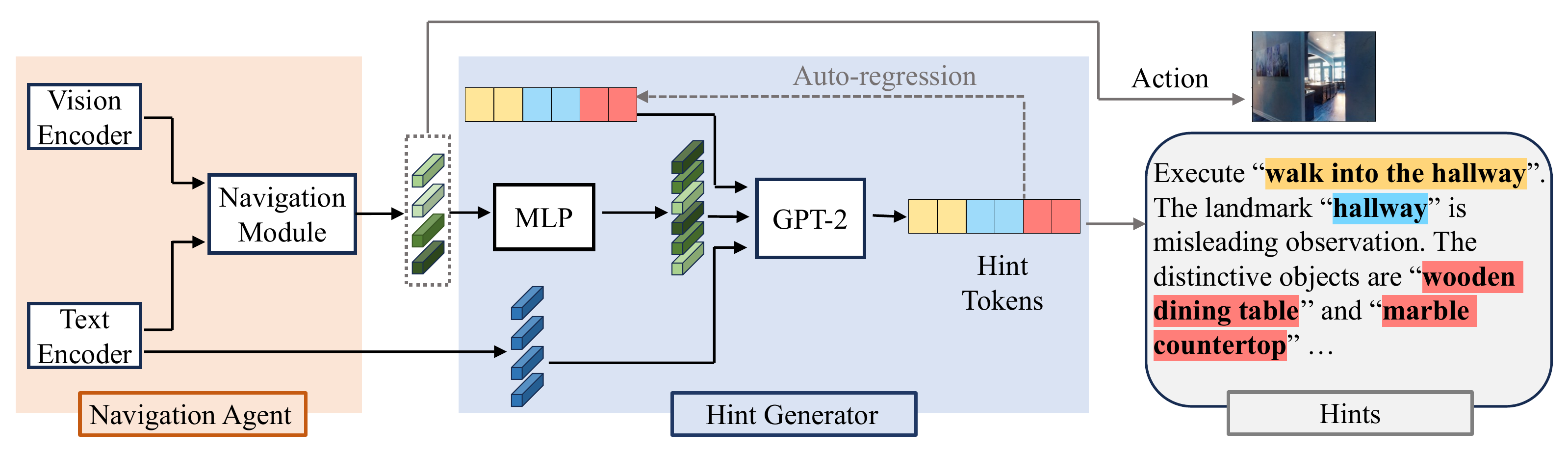}
    \caption{Model Architecture. We introduce a hint generator designed to help the navigation agent acquire a deep understanding of the visual environment.
    The weighted vision representations (\textcolor{cambridgeblue}{\rule{0.4cm}{0.25cm}}), used as image prefix, and the instruction text representation, used as instruction prefix (\textcolor{denim}{\rule{0.4cm}{0.25cm}}), are input into a GPT2 decoder. The decoder generates hints during navigation at each step.
    The hints include the three parts of sub-instruction (\textcolor{chromeyellow}{\rule{0.4cm}{0.25cm}}), landmark ambiguity (\textcolor{capri}{\rule{0.4cm}{0.25cm}}), and target distinctive objects (\textcolor{carminepink}{\rule{0.4cm}{0.25cm}}). \vspace{-5mm}}
    \label{fig:Architecture}
\end{figure*}

\subsection{VLN Agent with a Hint Generator}
We propose a hint generator that can be plugged into any navigation agent easily.
We use VLN$\circlearrowright$BERT~\cite{hong2020recurrent} as the base model to illustrate our method but noted that the hint generator is compatible with most of the current agents.
Fig.~\ref{fig:Architecture} shows the model architecture.\\
\noindent\textbf{Text Encoder} We use BERT~\cite{vaswani2017attention} to obtain initial text representation of instruction, denoted as $X = [x_1, x_2, \cdots, x_l]$. \\
\noindent\textbf{Vision Encoder} We follow previous works to concatenate image and relative orientation features as vision features for each candidate viewpoint. Specifically, we extract the image features from ResNet-152~\cite{he2016deep} % that is 
pre-trained on the Places365 dataset~\cite{zhou2017places}.
The orientation features are derived from the relative heading denoted as $\alpha$ and the elevation denoted as $\beta$. The orientation features are represented as $[\sin\alpha; \cos\alpha; \sin\beta; \cos \beta]$. 
The vision features are then passed through an MLP~(Multilayer Perception) of Vision Encoder to obtain vision representation for each candidate viewpoint, denoted as $[v_1, v_2, \cdots, v_n]$.\\
\noindent\textbf{Navigation Agent}
VLN$\circlearrowright$BERT is a cross-modal Transformer model. Besides text and vision representations, a state representation is introduced in the model to store history information recurrently, which is denoted as $S$. 
At the $t$-th navigation step, the text representation $X$, the visual representation $V_t$ and state representation  $S_t$ are input into cross-modal Transformer layers, as follows,
\begin{equation}
    \hat{X}, \hat{S_t}, \hat{V_t} = Cross\_Attn(X, [S_t;V_t]),
\end{equation}
where $\hat{X}$, $\hat{S_t}$, and $\hat{V_t}$ are the learnt contextual text, state representation, and visual representations, respectively.
Then we apply attention layer between state representation $\hat{S_t}$ contextual vision representations $\hat{V_t}$ as follows,
\begin{equation}
    S_{t+1}, a_t = Attn(k=\hat{V_t}, q=\hat{S_t}, v=\hat{V_t}),
\end{equation}
where $S_{t+1}$ is the updated state representation that is passed to the next steps to convey the history. $a_t$ is the attention score over the navigable views and serves as the action probability of the current step.

%, and it is the attention score between state representation $\hat{S_t}$ and visual representation $\hat{V_t}$ after self-attention.

\noindent\textbf{Hint Generator} Inspired by the idea of prefix engineering~\cite{mokady2021clipcap} that uses the image representation as the prefix of the text for the image captioning task,
we employ a decoder language model~(LM) and use the contextual visual representation of the navigation agent and the original instruction as the prefix. However, unlike the previous work, rather than just using one image as the prefix, we input all images of candidate viewpoints to encourage the hint generator to learn the global relations among views.

% The module will enforce the relevant representation in generating the explanation with grounding ability, improving the agent's navigation ability.
Formally, we denote the hint at the $i$-th navigation step as $C^i=\{c^i_1, c^i_2, \cdots, c^i_j\}$, where $j$ is the length of the hint.
Different from LANA~\cite{wang2023lana} that generates route description after navigation, our hint generator provides a more in-depth visual description at each step. Our approach requires the agent to possess a global and deep visual understanding, which can be learnt through the supervision from our navigation hint dataset explained in Section~\ref{hintdataset}.
We obtain the LM representation of the original instruction $W$ and the hint $C$ as $X'=\{x'_1, x'_2, \cdots, x'_l\}$ and  $c=\{c_1, c_2, \cdots, c_j\}$ respectively. 
Since the semantic structure of our auto-generated dataset can be easily captured, we use a 1.5B-parameters decoder LM~(GPT-2 large) in the hint generator.
Note that any larger decoder language model in the GPT series can be employed.

We use the instruction text representation $X'$ as the instruction prefix representation. We use the weighted vision representations output from the navigation agent as the image prefix representation. 
The weighted vision representation is obtained using action probability and the contextual vision representations as $\hat{\hat{V_t}} = a_t*\hat{V_t}$.
Then we simply employ an MLP to map $\hat{\hat{V_t}}$ to LM token space. We denote such MLP as $F$. We obtain prefix embedding that is mapped from visual representation $\hat{V}$ as follows,
\begin{equation}
    p_1, \cdots, p_k = F(\hat{\hat{V_t}}),
\end{equation}
where $k$ is the prefix length, and $p$ is the image prefix representation. 
We concatenate the representation of image prefix $p$ and instruction prefix $X'$, and combine them with the text representation of hint $C$. The hint generator only decodes the hint in an auto-regressive manner at each step.
During training, the parameters of both of MLP and the LM in the hint generator and the navigator are updated. The training objective is to maximize the likelihood of the next hint token. The following equation shows the loss of generating the $j$-th token of the hint at the $i$-th step.
%predict the next explanation tokens in an auto-regression manner.
%conditioned on the visual representations and the instructions in an autoregressive fashion.
\begin{equation}
\begin{aligned}
L_{hint}=-\sum_{i,j} \log p_\theta(c_j^i |
p^i_1, \cdots, p^i_k, \\
x'_1, \cdots, x'_l,  c^i_j, \cdots, c^i_{j-1}).
\end{aligned}
\end{equation}

 % It helps to the navigation agent learn he relevant representation by encourage the agent generate relevant explanations.
\begin{table*}[ht]
%\scriptsize
\small
%\footnotesize
    \begin{center}
    \resizebox{0.95\textwidth}{!}{
    \begin{tabular}{c c c c c c c c c c}
    \hline
      &  & \multicolumn{5}{c}{\textbf{Validation Unseen}}  & \multicolumn{3}{c}{\textbf{Test Unseen}}\\
    \hline
        %  & Validation-Unseen & Validation-Seen & Test(unseen)\\
        & Method & \textbf{NE} $\downarrow$ &  \textbf{SR} $\uparrow$ & \textbf{SPL}$\uparrow$ & \textbf{sDTW}$\uparrow$ & \textbf{nDTW}$\uparrow$ & \textbf{NE} $\downarrow$&
         \textbf{SR} $\uparrow$& \textbf{SPL} $\uparrow$\\
    \hline
    $1$ & Seq-to-Seq~\cite{anderson2018vision} & $7.81$ & $0.22$ & $-$ & $-$ & $-$ & $7.85$ & $0.20$ & $0.18$\\
    % $2$ & Speaker-follower~\cite{fried2018speaker} & $6.62$ & $0.36$ & $-$ & $-$ & $-$ & $6.62$ & $0.35$ & $0.28$ \\
    $2$ & Self-Monitor~\cite{ma2019self} & $5.52$ & $0.45$ & $0.32$ & $-$ & $-$ & $5.67$ & $0.48$ & $0.35$\\
    $3$ & AuxRN~\cite{ma2019self} & $5.63$ & $0.51$ & $0.46$ & $-$ & $-$ & $-$ & $-$ & $-$\\
    \hline
    $4$ &  VLN$\circlearrowright$BERT~\cite{hong2020recurrent} & $3.93$ & $0.63$ & $0.57$ & $-$ & $-$ & $4.09$ & $0.63$ & $0.57$ \\
    $5$ &  HAMT (ViT)~\cite{chen2021history} & $3.97$ & $0.66$ & $0.61$ & $-$ & $-$ & $3.93$ & $0.65$ & $\mathbf{0.60}$\\
     $6$ & LANA~\cite{wang2023lana} & $-$ & $0.66$ & $0.60$ & $-$ & $-$ & $-$ & $0.64$ & $0.59$\\
    $7$ &  VLN-SIG (ViT)~\cite{li2023improving} & $3.37$ & $0.68$ & $0.62$ & $0.59$ & $0.70$ & $-$ & $0.65$ & $\mathbf{0.60}$\\
    $8$ &  VLN-trans~\cite{zhang2023vln} & $3.34$ & $\mathbf{0.69}$ & $0.63$ & $0.60$ & $0.70$ & $3.94$ & $\mathbf{0.66}$ & $\mathbf{0.60}$\\
    \hline
    $9$ & EDrop$^{*}$~\cite{tan2019learning} & $5.49$ & $0.55$ & $0.47$ & $0.42$ & $0.58$ & $5.60$ & $0.51$ & $0.49$\\
    $10$ & EDrop + Hint. (NavHint) & $5.44$ & $0.55$ & $0.47$ & $0.44$   & $ 0.60$  & $5.47$ & $0.53$ & $0.49$\\
     \hline
    $11$ &  VLN$\circlearrowright$BERT$^{++}$ ~\cite{zhang2023vln}& $3.40$ & $0.67$ & $0.61$ & $0.58$ & $0.69$&  $4.02$ & $0.63$ & $0.58$\\
$12$&VLN$\circlearrowright$BERT$^{++}$ + Hint. (NavHint) & $\mathbf{3.23}$ & $\mathbf{0.69}$ & $\mathbf{0.65}$ & $\mathbf{0.61}$  & $\mathbf{0.72}$ & $4.00$ & $0.65$ & $\mathbf{0.60}$ \\
    \hline
    \end{tabular}
    }
    \end{center}
    \caption{Experimental results on R2R dataset. The best results are in bold font. VLN$\circlearrowright$BERT$^{++}$ is the improved version of VLN$\circlearrowright$BERT by pre-training the cross representations using a larger dataset~(see Sec~\ref{Implementation Details}). ViT: uses Vision Transformer representations. Hint.: uses our hint generator. \vspace{-2mm}}
    \label{R2R Experimental Result}
\end{table*}
\noindent\textbf{Training and Inference for the VLN Agent}
For the navigation, we train the navigation with a mixture of Imitation Learning~(IL) and Reinforcement Learning~(RL)~\cite{tan2019learning}. It consists of the cross-entropy loss of the predicted probability distribution against the ground-truth action and a sampled action from the predicted distribution to learn the designed rewards. In summary, the navigation loss is as follows,
\begin{equation}
\label{navigation objective}
\small
    L_{nav} = -\sum_t-\alpha^*_tlog(p^\alpha_t)-\lambda\sum_t\alpha^s_tlog(p^\alpha_t),
\end{equation}
where $\lambda$ is the hyperparameter to balance the two components, $\alpha^*_t$ is the teacher action for IL, and $\alpha^s_t$ is sample action for RL. 
We jointly train the navigation agent with hint generator using the following objective, 
\begin{equation}
    L = L_{hint} +L_{nav}.
\end{equation}
During inference of navigation, we use greedy search to select an action with the highest probability at each navigation step to generate a trajectory. To generate hint, we utilize the trained weighted visual representation and the original instruction text representation as prompts and employ a greedy search approach to generate the hints.
\section{Experiment}
\subsection{Dataset and Evaluation Metrics}
\textbf{Dataset} We evaluate our approach on R2R~\cite{anderson2018vision} and R4R datasets~\cite{jain2019stay}, which are built upon  Matterport3D simulator~\cite{anderson2018vision}. R2R includes $21,567$ instructions and $7,198$ trajectories. 
% The dataset has been partitioned into four sets: train, validation seen, validation unseen, and test unseen sets.
R4R is an extension of R2R to combine the two adjacent tail-to-head trajectories in R2R.
% It contains three sets: train, validation seen, and validation unseen. 
The visual environments in unseen sets are excluded in the training sets.
\\
\textbf{Evaluation Metrics} Three main metrics are used to evaluate navigation wayfinding performance~\cite{anderson2018vision}. (1) Navigation Error (NE) (2) Success Rate (SR) (3) Success Rate Weighted Path Length (SPL).
Another three metrics measure the fidelity between the predicted and the ground-truth trajectories. (4) Coverage Weighted by Length Score (CLS)~\cite{jain2019stay} (5) normalized Dynamic Time Warping (nDTW)~\cite{ilharco2019general} (6) Normalized Dynamic Time Warping weighted by Success Rate (sDTW). More details are in Appendix~\ref{appendix:dataset} and~\ref{appendix:evaluation metrics}.
% We provide more details about the dataset and metrics in Appendix~\ref{appendix:dataset} and Appendix~\ref{appendix:evaluation metrics}.

\subsection{Implementation Details}
\label{Implementation Details}
% We follow VLN$\circlearrowright$BERT$^{++}$~\cite{zhang2023vln} using their pre-trained weights to initialize our navigation model. 
We use pre-trained VLN$\circlearrowright$BERT$^{++}$~\cite{zhang2023vln} to initialize our navigation model. 
VLN$\circlearrowright$BERT$^{++}$ further trains the pre-trained weights in VLN$\circlearrowright$BERT~\cite{hong2020recurrent, hao2020towards} on a large image-text-action dataset including RXR~\cite{ku2020room}, Marky-mT5~\cite{wang2022less}, and SyFis~\cite{zhang2023vln}.
The dimensions of both BERT and GPT text representations are $768$-d. 
% For training the navigation agent with our hint generator, 
In the training, we conducted 300K iterations on an NVIDIA RTX GPU~($20$ hours), with a batch size of 8 and a learning rate of $1e-5$. $\lambda$ in Eq.~\ref{navigation objective} is $0.2$.
We set the maximum prefix length for each image as $10$ for the hint generator and the number of generated tokens as $80$. The best model is selected according to
performance on val unseen split. Please check our code~\footnote{https://github.com/HLR/NavHint.git} for the implementation.
% In terms of the hint generator, we set the maximum prefix length for each image as $10$. During inference, the maximum number of generated tokens is $80$. 

\subsection{Experimental Results}
Table~\ref{R2R Experimental Result} shows the performance on validation unseen and test of the R2R dataset in a \textit{single-run setting} where the navigation agent traverses without \textit{ backtracking} and \textit{pre-exploring}.
To verify the adaptability of our approach, 
we evaluate it using both LSTM-based and Transformer-based navigation agents.
Since Transformer-based methods are pre-trained on large vision-language datasets and have a more complex model architecture, they achieve a higher performance than LSTM-based methods.
For the LSTM-based model, we use EDrop~\cite{tan2019learning} % as our baseline. It
which uses CLIP~\cite{radford2021learning} visual representations without augmented data during training. For the Transformer-based model, we use the VLN$\circlearrowright$BERT$^{++}$ (row\#11) as the baseline.

Row\#1 to row\#3 in Table~\ref{R2R Experimental Result} show other LSTM-based methods and row\#4 to row\#8 are the SOTA Transformer-based methods.
Row\#9 shows the performance of the LSTM baseline EDrop.
Row\#10 shows the results after equipping the EDrop with our designed hint generator. The improved sDTW and nDTW on the validation unseen proves that the hint generator helps the navigation agent follow the instructions.
Moreover, our hint generator on top of the VLN$\circlearrowright$BERT$^{++}$ (row\#12) significantly improves both wayfinding metrics (SP and SPL) and fidelity metrics (sDTW and nDTW) of the baseline model, indicating that our hint generator not only assists the agent in reaching the correct destination but also encourages the agent to follow the original instructions. 
% LANA (row\#6) is to generate the original instructions during navigation; however, our significant improvement in validation unseen indicates the effectiveness of our explanation dataset.
Improving both LSTM-based and Transformer-based navigation agents shows the generalization ability of the navigation agent with our designed hint generator.

Table~\ref{Experiment Results for R4R} shows the results on the unseen validation of the R4R dataset.
We use VLN$\circlearrowright$BERT$^{++}$ as our baseline model (row\#7). 
Row\#1 to row\#3 are using LSTM-model, and row\#4 to row\#6 are using Transformer-based models.
The result of our method (row\#8) shows that we can improve SPL, sDTW, and CLS, that is, improving both the wayfinding and fidelity of the baseline models. These results are consistent with the improvements on the R2R dataset.
Though the VLN-Trans (row\#6) (SOTA) is very competitive, we additionally provide hints that can be used for explicitly analyzing the agent's decisions instead of implicit sub-instruction learning designed in VLN-Trans.

\begin{table}[t]
\small
    \begin{center}
    \resizebox{0.48\textwidth}{!}{
    \begin{tabular}{c c c c c c c}
    \hline
    & Method &  \textbf{NE}$\downarrow$ & \textbf{SR}$\uparrow$ & \textbf{SPL}$\uparrow$ & \textbf{CLS}$\uparrow$ & \textbf{sDTW}$\uparrow$  \\
    \hline
    $1$ & OAAM~\cite{qi2020object} & $13.80$  & $0.29$ & $0.18$ & $0.34$  & $0.11$\\
    $2$ & RelGraph~\cite{hong2020language} & $7.55$ & $0.35$ & $0.25$ & $0.37$  & $0.18$\\
    $3$ & NvEM~\cite{an2021neighbor} & $6.80$ & $0.38$ & $0.28$ & $0.41$ & $0.20$\\
    \hline
    $4$ & VLN$\circlearrowright$BERT~\cite{hong2020recurrent}  & $6.48$ & $0.43$ & $0.32$ & $0.42$ & $0.21$\\
    $5$ & CITL~\cite{liang2022contrastive} & $6.42$ & $0.44$ & $0.35$ & $0.39$ & $0.23$ \\
    $6$ & VLN-Trans~\cite{zhang2023vln}  &  $\mathbf{5.87}$ & $\mathbf{0.46}$ & $\mathbf{0.36}$ & $\mathbf{0.45}$ & $\mathbf{0.25}$\\
    \hline
    $7$ & VLN$\circlearrowright$BERT$^{++}$ ~\cite{zhang2023vln} & $6.33$ & $0.44$ & $0.34$ & $0.43$ & $0.23$\\
    $8$ &VLN$\circlearrowright$BERT$^{++}$ + Hint. (NavHint) &  $6.04$ & $\mathbf{0.46}$ & $\mathbf{0.36}$ & $\mathbf{0.45}$ & $\mathbf{0.25}$\\
    \hline
    \end{tabular}
    }
    \end{center}
    \vspace{-3mm}
    \caption{\small Results on R4R validation unseen dataset.\vspace{-5mm}}
    \label{Experiment Results for R4R}
\end{table}

\begin{table}[t]
    \centering
    \resizebox{0.48\textwidth}{!}{
    \begin{tabular}{c c c c c}
    \hline
          Model& \multicolumn{2}{c}{\textbf{Val Seen}} & \multicolumn{2}{c}{\textbf{Val Unseen}} \\
          & \texttt{Bleu}-1 & \texttt{Bleu}-4 & \texttt{Bleu}-1 & \texttt{Bleu}-4 \\
         \hline
         % EDrop-speaker~\cite{tan2019learning} & - & $0.25$ & - & $0.24$ \\
         EDrop + Hint. (ours) & $0.74$ & $0.62$ & $0.72$ & $0.60$\\
         % LANA~\cite{wang2023lana} & $0.76$ & $0.31$ & $0.74$ & $0.30$ \\
        VLN$\circlearrowright$BERT$^{++}$+ Hint. (ours) & $\mathbf{0.76}$ & $\mathbf{0.64}$& $\mathbf{0.74}$  & $\mathbf{0.62}$ \\
         \hline
    \end{tabular}
    }
    \caption{\texttt{Bleu} score for the generated sub-instruction on the R2R dataset.\vspace{-4mm}}
    \label{tab:bleu}
\end{table}

\begin{table}[t]
%\scriptsize%\small%\footnotesize
\small
\renewcommand{\tabcolsep}{0.35em}
    \begin{center}
    \resizebox{0.49\textwidth}{!}{
    \begin{tabular}{c|cccc|ccc}
    \hline
    {\textbf{Method}}& \multicolumn{4}{c}{\textbf{Hints}} &  \multicolumn{3}{|c}{\textbf{Val Unseen}} \\
     & \textbf{Sub.} & \textbf{L-A.} & \textbf{TD-Obj.} & \textbf{Obj.} & \textbf{SR}$\uparrow$ & \textbf{SPL}$\uparrow$ &\textbf{nDTW}$\uparrow$ \\
    \hline
     Baseline & & & & & $0.665$ & $0.607$ & $0.685$  \\
    \hline
    1 & \ding{52} && && $0.671$ & $0.612$ & $0.690$ \\
    2 & & \ding{52} & & & $0.673$ & $0.613$ & $0.687$ \\
    3 & & & \ding{52} & & $0.677$ & $0.624$ & $0.702$\\
    4 & & & & \ding{52} & $0.676$ & $0.621$ & $0.698$ \\
    \hline
    5 &  \ding{52} &  \ding{52}& &  & $0.674$ & $0.614$ &  $0.709$ \\
    6 &  \ding{52} &\ding{52} & & \ding{52} &  $0.681$ & $0.632$ & $0.694$ \\
    7 & \ding{52}& \ding{52} & \ding{52}& & $\mathbf{0.692}$ & $\mathbf{0.647}$ & $\mathbf{0.724}$  \\
    \hline
    \end{tabular}
    }
    \end{center}
    \vspace{-3mm}
    \caption{\small Ablation study, where Baseline is VLN$\circlearrowright$BERT$^{++}$. \small Sub.:sub-instruction; L-A.:Landmark Ambiguity; TD-Obj: Target Distinctive Objects. Obj:Top-3 objects.\vspace{-4.5mm}}
    \label{Ablation study}
\end{table}

\subsection{Ablation Study}
% We conduct an ablation study to assess the impact of each part of the hint on the agent's navigation performance.
% Table~\ref{Ablation study} shows the results.
Table~\ref{Ablation study} reports the ablation analysis.
From row\#1 to row\#3, we individually include sub-instruction, landmark ambiguity, and targeted distinctive objects to the hint. All navigation performance metrics improve gradually compared to the baseline.
In another experiment (row\#4), we attempt to describe the visual environment by identifying only top-3 recognized objects (using MiniGPT-4) in the targeted viewpoint without differing them from other viewpoints. The navigation results still improve, indicating that visual descriptions of the objects benefit the overall navigation performance.
Row\#5 shows that combining sub-instruction and landmark ambiguity further improves the baseline, particularly in the nDTW metric.
In row\#6, when we combine sub-instruction, landmark ambiguity and top-3 objects, we observe improvement in the goal-related metrics (SR and SPL), but the model's ability to faithfully follow the instruction is somewhat compromised (lower nDTW).
The best result is obtained when we replace the above top-3 objects with distinctive ones (row\#7), indicating our designed hint's effectiveness in describing the targeted view from a global perspective.

% This means the model learns the navigation rational behind this. Also, sub-instruction, landmark ambiguity and non distinctive also helps the romance, but not that much as distinctive objects, which further indicates its effectiveness.

% the result is around baseline. Then we combine two of 
% Intuitively, it is interesting to see maybe just let the agent describe the objects in the visual environment is already enough. Then we conduct an experiment just generate the top3 visible objects in the target view. However, the result is almost same the baseline. This means only talk about the objects may not influence the performance that much. 

%  that the performance drops when we remove each part of the explanation.
% It is intuitively expected that removing sub-instruction descriptions (row\#2) and target distinctive object descriptions (row\#4) from the explanation would decrease navigation performance. This is because these components provide supervision to the model to identify the progress and understand the environment in depth for generating correct distinctive explanations.
% Moreover, when we exclude the description of landmark ambiguity~(row\#3), we observe a notable decrease of approximately $3\%$ in both the seen and unseen datasets. 
% This result emphasizes that boosting the grounding ability of textual landmarks into visual objects is essential for the VLN task.
% , as it enhances the grounding ability from textual landmarks to visual objects.

\subsection{Generated Hints Analysis}
In this section, we assess the content of each part of the generated hints on the R2R validation dataset to analyze agent's grounding ability.

\noindent\textbf{Sub-instruction Analysis}
We use \texttt{Bleu} score~\cite{papineni2002bleu} as an evaluation metric to assess whether the navigation agent can identify sub-instruction accurately. We conduct experiments on both LSTM-based and Transformer-based navigation agents, as shown in Table~\ref{tab:bleu}. 
The generated sub-instruction from the Transformer-based navigation agent can obtain a relatively high \texttt{Bleu} score compared to the LSTM-based agent. This result demonstrates that a more robust navigation agent achieves a stronger alignment between the instruction and visual modality for identifying the relevant part of the instruction to track the progress.
% that score of our method demonstrates the agent's strongalignment and grounding between the instruction and visual modality foridentifying the relevant part of the instruction at each step and tracking the progress.

% Since there is no other method generating sub-instruction as us, we provide our comparison with 
% Note that the compared methods generate full route descriptions, whereas our approach generates sub-instructions at each step. The relatively high \texttt{Bleu}-4 score of our method demonstrates the agent's strongalignment and grounding between the instruction and visual modality foridentifying the relevant part of the instruction at each step and tracking the progress. . \pk{it is not clear to me how you compare to Lana and if the comparison makes sense given the totally different generation settings}

\begin{figure}
    \centering
    \begin{overpic}[width=1.0\linewidth]{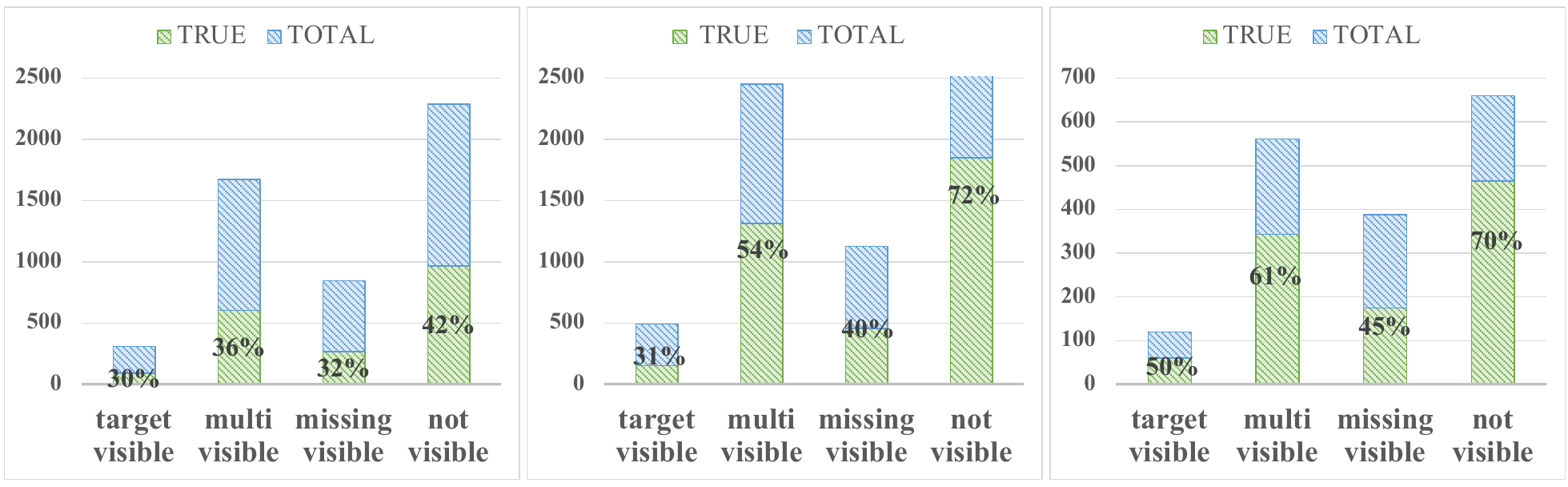}  
    %% caption
    \put(5,-3.5){\scriptsize{(a) EDrop+Hint.}}
    \put(33,-3.5){\scriptsize{(b) VLN$\circlearrowright$BERT$^{++}$+Hint.}}
    \put(75,-3.5){\scriptsize{(c) Correct Sub.}}
    \end{overpic}
    \caption{\small Accuracy of the generated landmark ambiguity. 
     Sub.: Sub-instruction.}
    \label{fig:sub_corr}
\end{figure}

% TOTAL: the total number of examples generating each category of landmark ambiguity. TRUE: the number of examples the landmark ambiguity is correct.
\noindent\textbf{Landmark Ambiguity Analysis}
We assess the accuracy of four categories of landmark ambiguity in the generated hints. Specifically, We extract the part of the landmark ambiguity from the generated hint and check its accuracy in the visual environment. 
% For example, suppose the generated hint is "The landmark table is observed in multiple views", we check if the object "table" occurs in multiple views in the visual environment. 
In Figure 4, the TOTAL in the y-axis shows the total number of navigation steps that include each ambiguity category, shown on the x-axis. The TRUE (green) indicates the percentage of navigation steps when the corresponding ambiguity truly exists. We evaluate both LSTM-based and Transformer-based agents, and the result shows that Transformer-based agents can achieve higher accuracy of landmark ambiguity. We conclude that accurate landmark ambiguity detection is positively correlated with better navigation performance. In Figure 4(c), we evaluate the generated hint for the examples in which the sub-instruction is generated correctly, as indicated by a \texttt{Bleu}-4 score of $1.0$. In those examples, the accuracy of identifying each category of landmark ambiguity is also higher. This result shows accurately locating the sub-instruction positively impacts landmark ambiguity detection.
% we evaluate the generated landmark ambiguity and environment processing to measure the accuracy of the type of ambiguity generated by the model. 
% Fig.~\ref{fig:sub_corr} displays the number of examples of generating each ambiguity category using our LSTM-based (Fig. 4(a)) and Transformer-based (Fig. 4(b)) methods.

%The result shows that our Transformer-based agent achieves higher accuracy regarding landmark ambiguity clarification.
% The result shows that the Transformer-based agent achieves higher accuracy, suggesting that accurate landmark ambiguity clarification is positively correlated to better navigation performance.
% Besides, 
% there is a strong relationship between landmark ambiguity and the generation of correct sub-instructions. 
% we examine the cases where good sub-instructions are generated, indicated by a \texttt{Bleu}-4 score of $1.0$, and analyze their impact on the ambiguity identification accuracy. As depicted in Fig. 4(c), the accuracy improves in all categories when the sub-instructions are correct, highlighting the significance of accurately locating the navigation progress.

\begin{figure}
    \centering
    \begin{overpic}[width=1.0\linewidth]{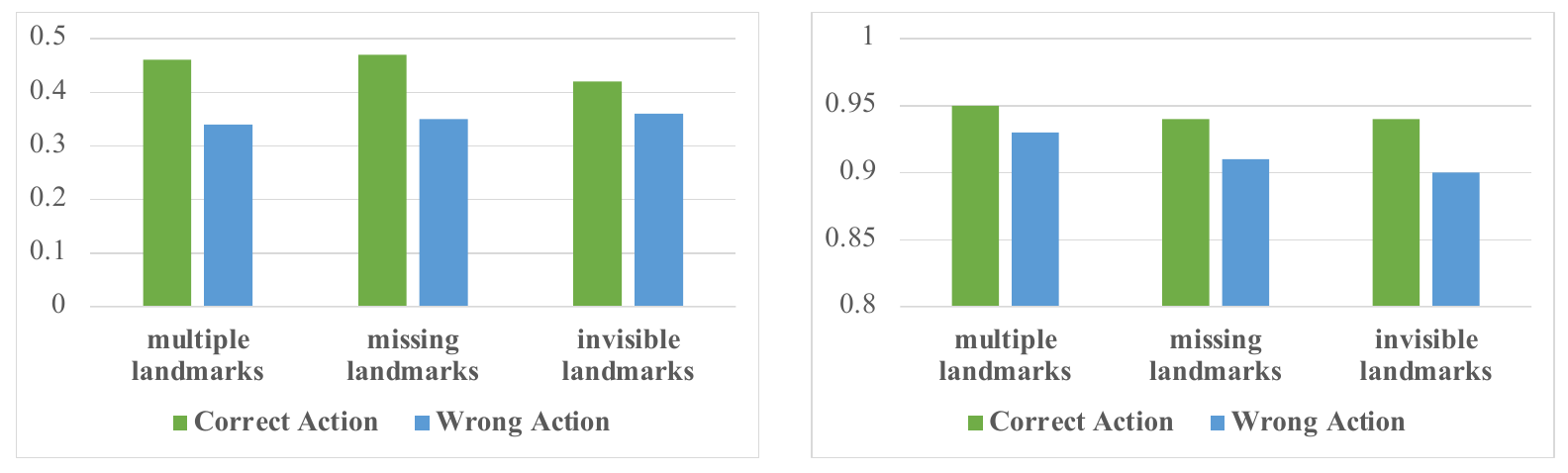}  
    %% caption
    \put(10,-3.5){\scriptsize{(a) Exact Matching}}
    \put(60,-3.5){\scriptsize{(b) Object Matching}}
    \end{overpic}
    \caption{\small Accuracy of the generated distinctive objects for each landmark ambiguity in the targeted viewpoint. \vspace{-5mm}}
    \label{fig:distinctive}
\end{figure}

\noindent\textbf{Targeted Distinctive Objects Analysis}
We report the accuracy of identifying the targeted distinctive objects in the generated hints when landmark ambiguity exists, as shown in Fig.~\ref{fig:distinctive}. The generated hints are from the model of VLN$\circlearrowright$BERT$^{++}$ with our designed hint generator. 
We provide two types of comparisons, exact phrase matching and object token matching while performing both wrong and right actions.
Exact matching evaluates the detection of distinctive object tokens and the attribute descriptions in the whole referring phrase. 
Object matching only evaluates the detection of distinctive object tokens.
The result shows that the accuracy in generating distinctive objects is generally higher when the action is correct than when it is wrong.
Also, the agent tends to generate distinctive objects that align with its targeted viewpoint, as indicated by an accuracy exceeding 90\%, even when the action is incorrect. 
The lower accuracy of exact matching also aligns with the fact that generating the whole referring expression, including the correct attributes, is more challenging.
% \begin{figure}
%     \centering
%     \includegraphics[height=5.5cm, width=1.0\linewidth]
%     {images/qualitative1.pdf}
%     \caption{Two qualitative examples. The green and orange arrows show the target and the predicted viewpoints. \vspace{-5mm}}
%     \label{fig:qualitative}
% \end{figure}

\begin{figure}[t]
\centering
\includegraphics[width=1.0\linewidth]{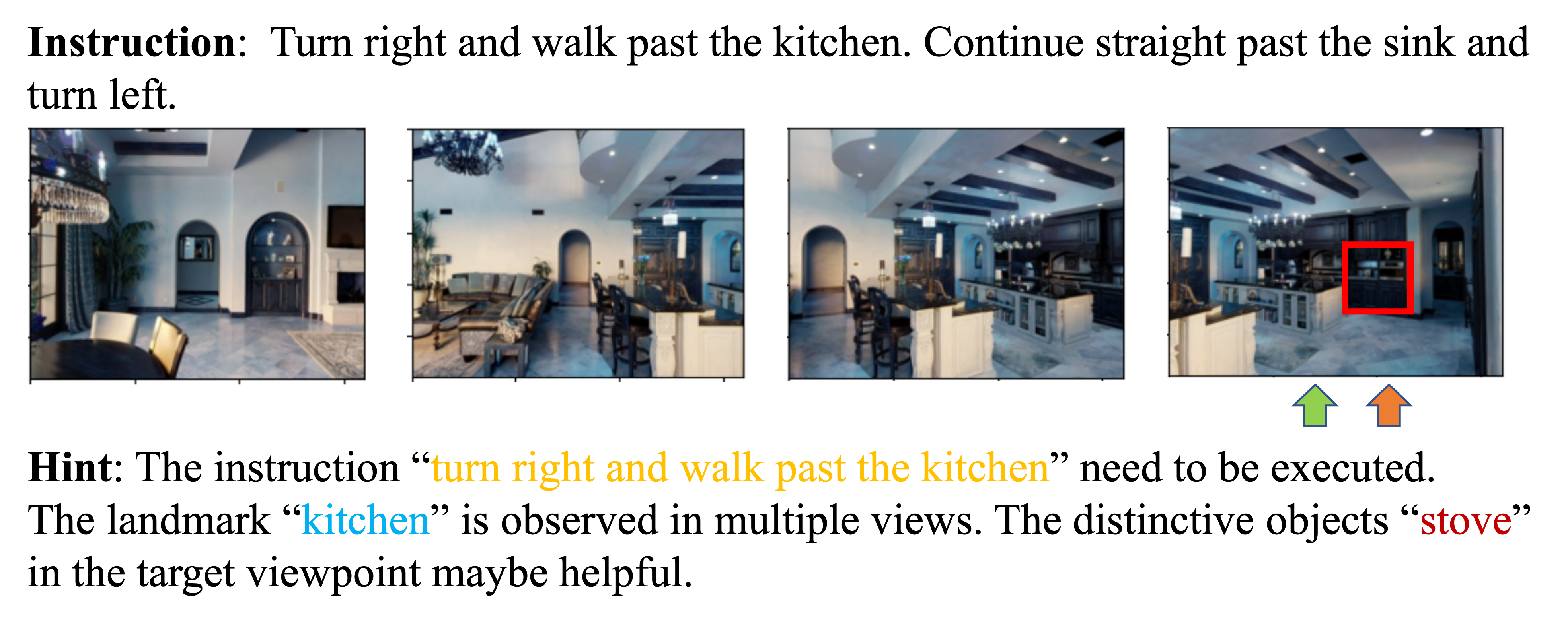}
\includegraphics[width=1.0\linewidth]{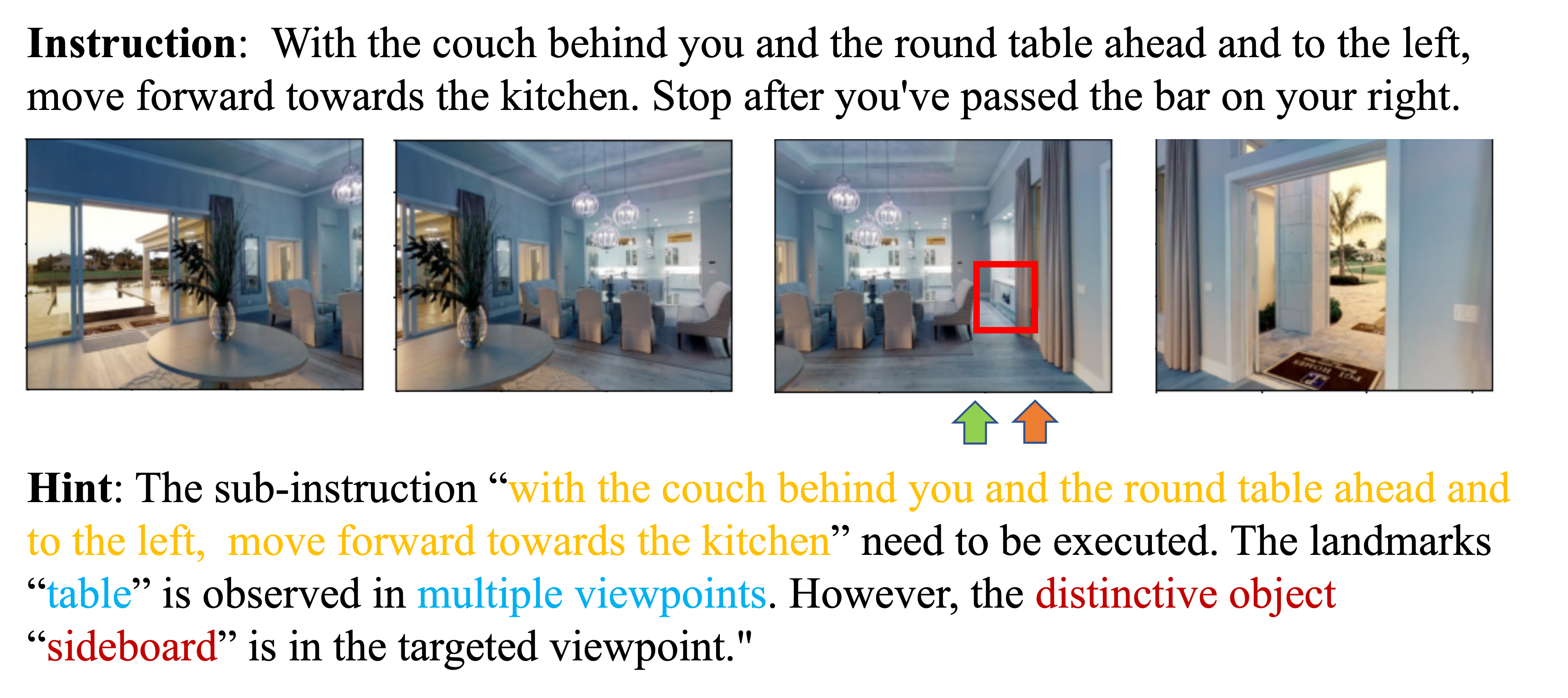}
\includegraphics[width=1.0\linewidth]{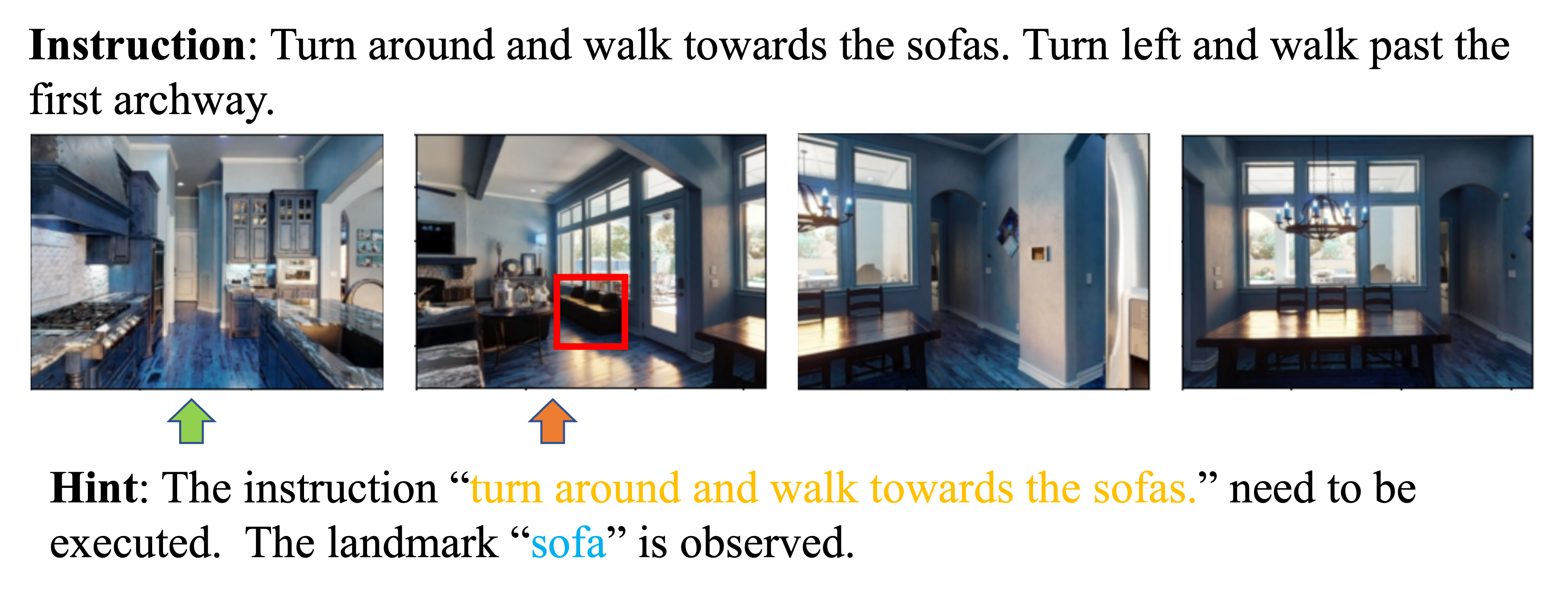}
\caption{Qualitative examples. The green and orange arrows show the ground-truth and the predicted viewpoints, respectively. \vspace{-5mm}}
\label{fig:qualitative}
\end{figure}

\subsection{Qualitative Examples}
% In Fig.~\ref{fig:qualitative}, we present two examples to illustrate the generated description by the navigation agent. 
Fig.~\ref{fig:qualitative} demonstrates a few examples of the generated descriptions.
The first two examples show successful cases where the agent makes a correct decision. The first example shows the agent can accurately identify the sub-instruction and notice the ambiguous landmark ``kitchen''. Then, it correctly pinpoints the distinctive object ``stove'', which only appears in the target viewpoint. In fact, our \textit{targeted distinctive object} design can help connect the specific object (e.g. stove, refrigerator, counter table) to more general scene objects (e.g. kitchen). Also, the second example shows the agent accurately points out the ``table'' in the instruction that appears in multiple viewpoints and refers to the ``sideboard'' in the target viewpoint. The third example shows a failure case in which the agent makes a wrong decision. % Although 
The sub-instruction is correctly identified, but the agent should turn around towards the counter table and proceed to the sofa rather than walk to the sofa directly. 
This further indicates that our descriptor pushes the model to focus on landmarks directly and ignore the directions and motions in the instruction. Despite this, our model can generate a description consistent with its selection.
More examples are in the Appendix~\ref{more qualitative example}. 

% Those examples show challenging scenarios  where the baseline model makes mistakes while our model can make the correct decisions. 
% Our model can provide the corresponding explanation, including the correct sub-instruction, landmark ambiguity, and distinctive objects in the selected viewpoint.

% \begin{table}[]
%     \centering
%     \resizebox{0.55\textwidth}{!}{
%     \begin{tabular}{c|c|c|c|c}
%     \hline
%         & multi visible& target visible & missing visible & not visible \\
%         multi visible & $49\%$ & $10\%$ & $14\%$ & $26\%$\\
%         target visible & $60\%$ & $4\%$ & $15\%$ & $21\%$\\
%         missing visible & $29\%$ & $3\%$ & $33\%$ & $35\%$\\
%         not visible & $18\%$ & $5\%$ & $14\%$ & $63\%$\\
%     \hline
%     \end{tabular}
%     }
%     \caption{Caption}
%     \label{tab:my_label}
% \end{table}

\section{Conclusion}
In this paper, we equip the navigation agent with a hint generator to generate visual descriptions during navigation, % enhancing 
which helps the agent's understanding of the visual environment. To train the hint generator, we create a navigation hint dataset that provides comprehensive supervision for training the agent. 
During navigation, the agent generates natural language descriptions about its visual environment at each step, including comparing various views and explaining ambiguities in recognizing the target destination.
% To train the navigation agent with the hint generator, we create a navigation hint dataset that provides supervision to guide the navigation agent to pinpoint a part of the sub-instruction that needs to be executed, understand the commonalities of multiple views in terms of shared landmarks that may cause ambiguity, and find the distinctive objects in the target view that differentiate it from other viewpoints.
% Our experimental results 
Empirical results show that detailed visual description generation improves both navigation performance and the interpretability of actions taken by the navigation agent.
% We also comprehensively analyze the agent's grounding ability by examining the generated hints. Our method not only improves the interpretability of the navigation agent but also provides a way for the agent to interact with humans effectively.

% can use our dataset to analyze the agent's grounding ability by examine the generated hint
% the grounding ability of the agent leading to improvements in both navigation performance as well as interpretability of the model and its communication with humans.
% We designed a corresponding  that can be plugged into any navigation agent \pk{to jointly perform actions and generate explanations at each step of navigation}.  \pk{This enhances its grounding ability and gaining a better understanding of the visual environment.: You can Combine this with the next sentence, start with " " } Experimental results demonstrate the effectiveness of our approach in improving both navigation performance and interpretability of the model.

\section{Limitations}
We mainly summarize the following limitations. First, although we employ the GPT2 language decoder, more recent and powerful GPT-series language decoders are now available and could be utilized. Exploring these advanced language decoders could potentially enhance the performance of our approach.
Second, we do not include more advanced vision representations, such as ViT representation, to train the navigation agent. We can surpass other methods using ResNet, but it would be interesting
to experiment with those different visual representations to generate better hints. 
Third, utilizing object visual descriptions from MiniGPT-4 may entail hallucination issues, which is a general challenge of VLMs. However, in our specific usage of MiniGPT4, we barely face this issue in the experiments.

% Third, our model is trained on a simulated environment that is split into discrete images, and this will have limitations when deployed in a real-world setting.

% Fourth, running our models needs access to powerful GPU resources that can limit easy accessibility for the public.
\section{Acknowledgement}
This project is supported by the National Science Foundation (NSF) CAREER award 2028626 and partially supported by the Office of Naval Research
(ONR) grant N00014-20-1-2005 and  grant N00014-23-1-2417. Any opinions,
findings, and conclusions or recommendations expressed in this material are those of the authors and
do not necessarily reflect the views of the National Science Foundation nor the Office of Naval Research. We thank all reviewers for their thoughtful
comments and suggestions.

% Entries for the entire Anthology, followed by custom entries
\bibliography{anthology,custom}

\clearpage
\appendix
\section{Appendix}

\subsection{Statistics of the VLN Hint Dataset}
\label{appendix: data statistics.}
We built VLN explanation dataset upon R2R dataset. We split our explanation dataset into train, validation seen, and validation unseen sets according to R2R.
We create explanation for each navigation step of trajectory given the corresponding instruction.
For train set, there are $4,675$ trajectories, and we create $69,969$ explanation in $61$ visual scenes. For validation seen set, there are $340$ trajectories, and we create $5,175$ explanations in $61$ visual scenes. For validation unseen set, there are $783$ trajectories, and we create $11,664$ explanations in $11$ visual scenes.

\subsection{Dataset}
\label{appendix:dataset}
We evaluate our approach on R2R~\cite{anderson2018vision} and R4R datasets~\cite{jain2019stay}, which are built upon  Matterport3D simulator~\cite{anderson2018vision}. R2R includes $21,567$ instructions and $7198$ trajectories. The dataset has been partitioned into four sets: train ($61$ scenes, $14,039$ instructions), validation seen ($61$ scenes, $1,021$ instructions), validation unseen ($11$ scenes, $2,349$ instructions), and test unseen sets ($18$ scenes, $4,173$ instructions).
R4R is an extension of R2R to combine the two adjacent tail-to-head trajectories in R2R. It contains three sets: train ($61$ scenes, $233,613$
instructions), validation seen ($61$ scenes, $1,035$ instructions), validation unseen ($11$ scenes, $45,162$ instructions). The scenes in unseen sets are not trained. \\

\subsection{Evaluation Metrics}
\label{appendix:evaluation metrics}
Three main metrics are used to evaluate navigation wayfinding performance~\cite{anderson2018vision}: (1) Navigation Error (NE): the mean of the shortest path distance between the agent's final position and the goal destination. (2) Success Rate (SR): the percentage of the predicted final position being within 3 meters from the goal destination. (3) Success Rate Weighted Path Length (SPL): normalizes success rate by trajectory length.
Another three metrics are used to measure the fidelity between the predicted and the ground-truth trajectory.
(4) Coverage Weighted by Length Score (CLS)~\cite{jain2019stay} (6) nDTW~~\cite{ilharco2019general}: Normalized Dynamic Time Warping: penalizes
deviations from the ground-truth trajectories. (6) Normalized Dynamic Time Warping weighted by Success Rate (sDTW)~\cite{ilharco2019general}: penalizes
deviations from the ground-truth trajectories and also considers the success rate.

\subsection{More Qualitative Examples}
\label{more qualitative example}
We present additional qualitative examples in this section. The first three are successful cases where the navigation agent makes correct actions, and the hint generator accurately generates sub-instruction, landmark ambiguity and distinctive objects in the instruction. The last two examples are failure cases. Despite incorrect actions, the agent still generates accurate distinctive objects within its selected viewpoint. The failures might come from inaccuracies in landmark extraction, which subsequently affect ambiguity checking.

\begin{figure}[t]
\includegraphics[width=1.0\linewidth]{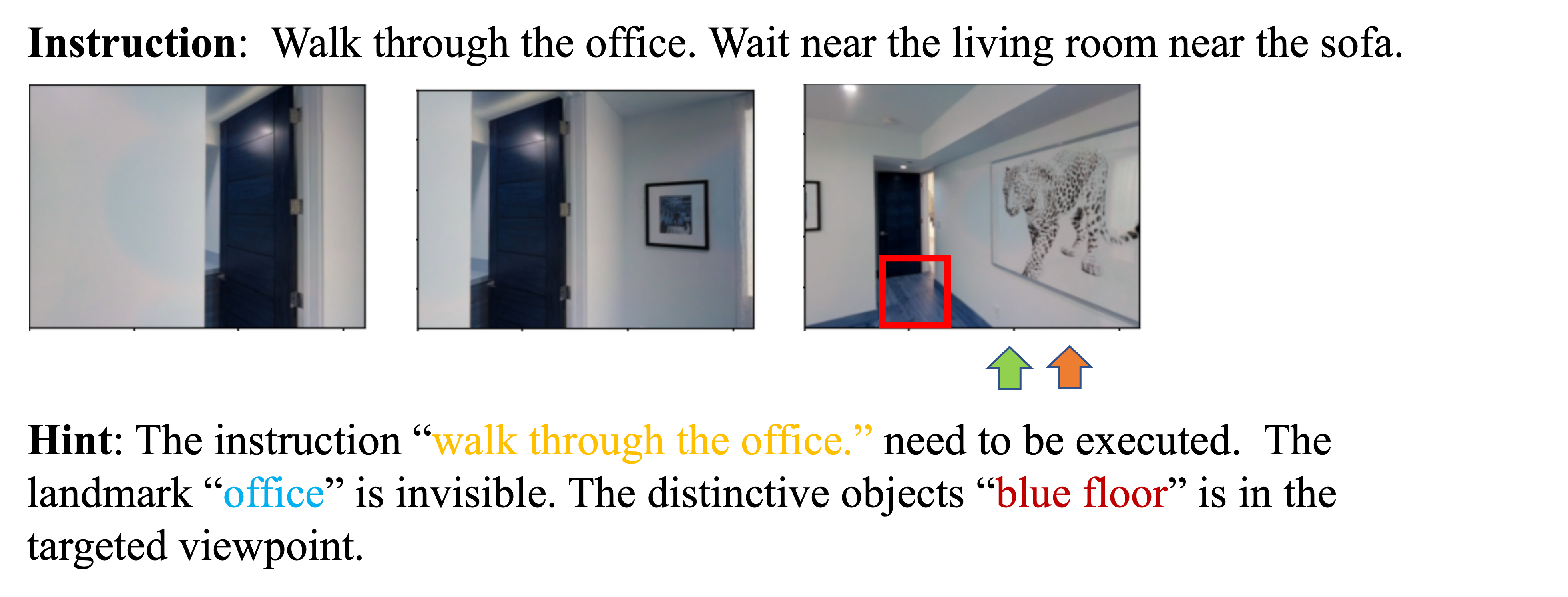}
\includegraphics[width=1.0\linewidth]{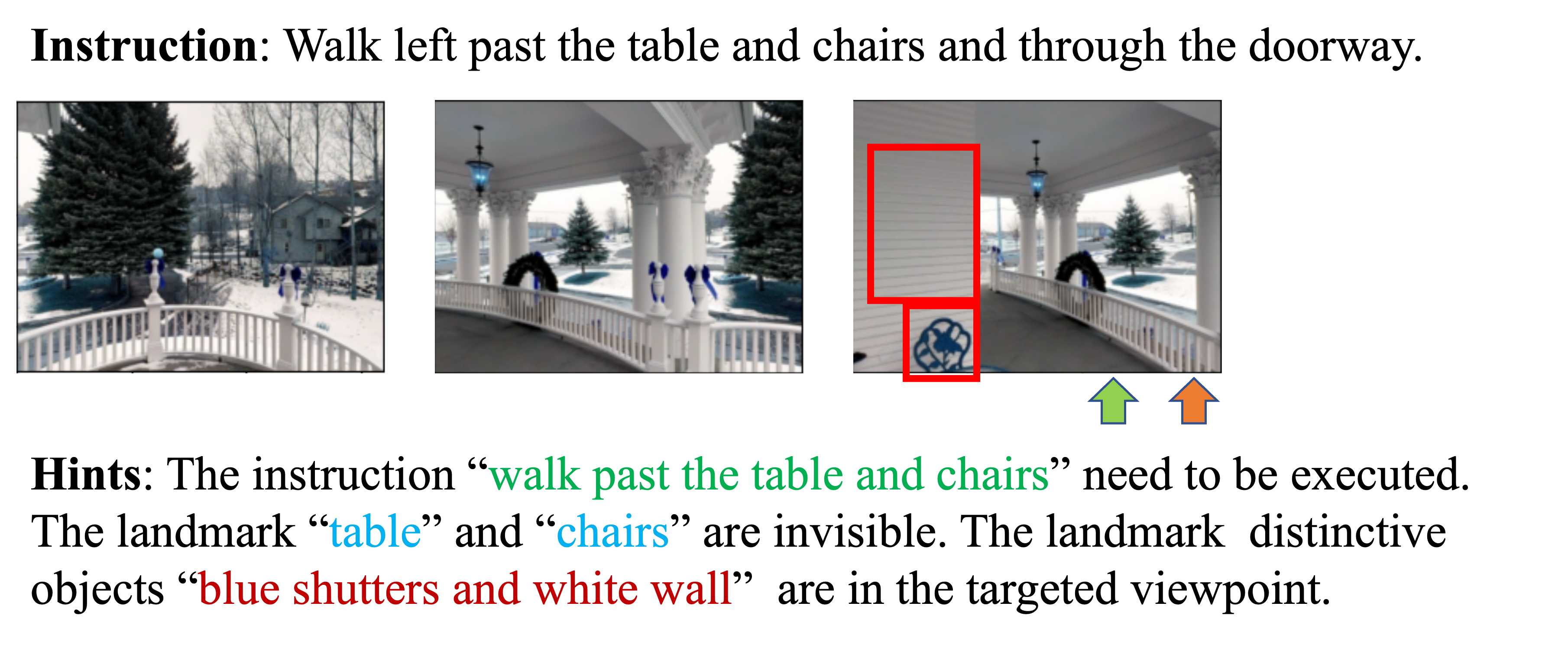}
\includegraphics[width=1.0\linewidth]{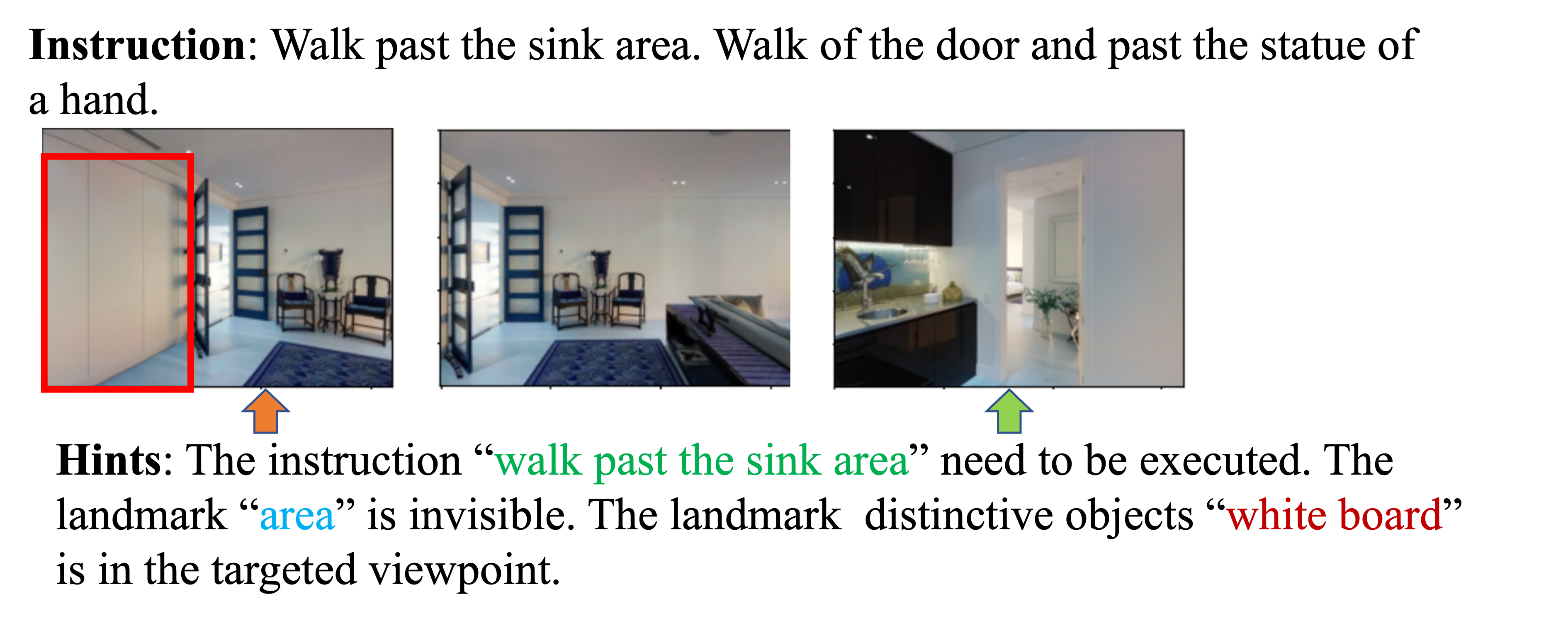}
\includegraphics[width=1.0\linewidth]{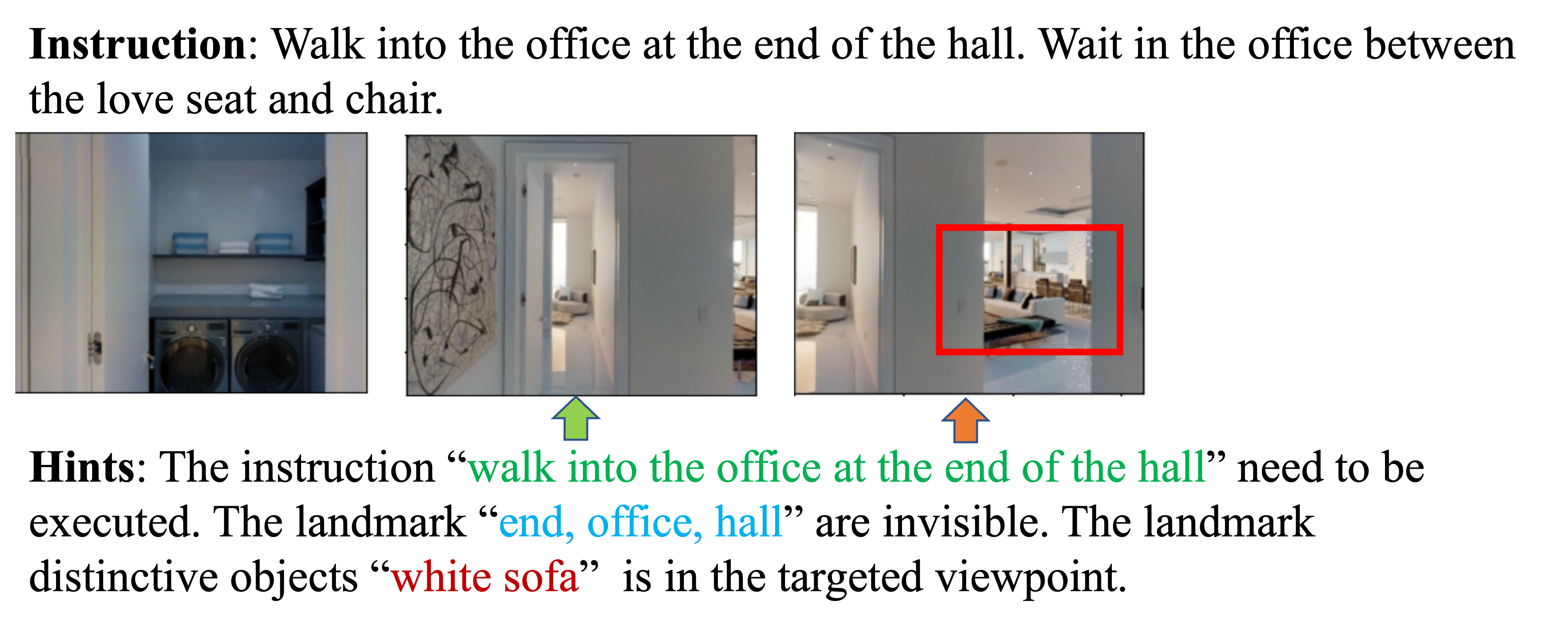}
\caption{More qualitative examples. The green and orange arrows show the ground-truth and the predicted viewpoints, respectively.}
\label{qualitative example2}
\end{figure}

% \begin{table}[]
%     \centering
%     \resizebox{0.5\textwidth}{!}{
%     \begin{tabular}{c|c|c|c|c}
%     \hline
%         & multi visible& target visible & missing visible & not visible \\
%         \hline
%         multi visible & $72\%$ & $4\%$ & $14\%$ & $10\%$\\
%         target visible & $57\%$ & $12\%$ & $2\%$ & $19\%$\\
%         missing visible & $27\%$ & $3\%$ & $39\%$ & $31\%$\\
%         not visible & $8\%$ & $4\%$ & $12\%$ & $76\%$\\
%     \hline
%     \end{tabular}
%     }
%     \caption{The performance of Challenge Clarification.}
%     \label{tab:challenge_case}
% \end{table}
% Table~\ref{tab:challenge_case} displays the percentage of examples where one challenge type is incorrectly labeled as other types. the model demonstrates good performance in correctly recognizing challenges. However, most examples of ``target visible" are mistakenly recognized as ``multi visible".

% \section{Example Appendix}
% \label{sec:appendix}

% This is a section in the appendix.

\end{document}